\pdfoutput=1

\documentclass[11pt,a4paper]{article}
\usepackage[hyperref]{acl2021}
\usepackage{times}
\usepackage{latexsym}

\usepackage{soul}

\usepackage{microtype}

\aclfinalcopy 

\newcommand{\spapi}[1]{{\textcolor{teal}{#1}}}

\title{Dealing with training and test segmentation mismatch:
FBK@IWSLT2021}

\author{Sara Papi\textsuperscript{1,2}, Marco Gaido\textsuperscript{1,2}, Matteo Negri\textsuperscript{1}, Marco Turchi\textsuperscript{1} \\
  \textsuperscript{1}Fondazione Bruno Kessler, Trento, Italy \\
  \textsuperscript{2}University of Trento, Italy \\
  \texttt{\{spapi|mgaido|negri|turchi\}@fbk.eu}}

\date{}

\begin{document}
\maketitle
\begin{abstract}
This paper describes FBK's system submission to the IWSLT 2021 Offline Speech Translation task. 
We participated with a direct model, which is a Transformer-based architecture trained to translate English speech audio data into German texts. The training pipeline is characterized by knowledge distillation and a two-step  fine-tuning procedure. Both knowledge distillation and the first fine-tuning step are carried out on manually segmented real and synthetic data, the latter being generated with an MT system trained on the available corpora. Differently, the second fine-tuning step is carried out on a random segmentation of the MuST-C v2 En-De dataset.
Its main goal is to reduce the performance drops occurring when a speech translation model trained on manually segmented data (i.e. an ideal, sentence-like segmentation) is evaluated on automatically segmented audio (i.e. actual, more realistic testing conditions).
For the same purpose, a custom hybrid segmentation procedure that accounts for both audio content (pauses) and for the length of the produced segments is applied to the test data before passing them to the system.
At inference time, we compared this procedure with a baseline segmentation method based on Voice Activity Detection (VAD). Our results indicate the effectiveness of the proposed  hybrid approach, shown by a reduction of the gap with manual segmentation from 8.3 to 1.4 BLEU points.
\end{abstract}

\section{Introduction}
\label{sec:intro}
Speech translation (ST) is the task of translating a speech uttered in one language into its textual representation in a different language. 
Unlike \textit{simultaneous} ST, where the audio is translated as soon as it is produced, in the \textit{offline} setting the audio is entirely available and translated at once.
In
continuity with the last two rounds of the IWSLT evaluation campaign \cite{iwslt_2019,ansari-etal-2020-findings},
the IWSLT2021 Offline Speech Translation task \cite{iwslt:2021}
focused on the translation into German of English audio data extracted from TED talks.
Participants could approach the task either with a cascade architecture or with a direct end-to-end system. The former represents the traditional pipeline approach \cite{StentifordSteer88,Waibel1991b} comprising an automatic speech recognition  (ASR) followed by a machine translation (MT) component. The latter \cite{berard_2016,weiss2017sequence} relies on a single neural network trained to translate the input audio into target language text bypassing any intermediate symbolic representation steps.

The two paradigms have advantages and disadvantages. 
Cascade architectures have historically guaranteed higher translation quality \cite{iwslt_2018,iwslt_2019} thanks to the large corpora available to train their ASR and MT sub-components. However, a well-known drawback of pipelined solutions is represented by error propagation: transcription errors are indeed hard (and sometimes impossible) to recover during the translation step. 
Direct models, although being 
penalized by the paucity of training data, have two theoretical competitive advantages, namely:  \textit{i)} the absence of error propagation as there are no intermediate 
processing 
steps, and \textit{ii)} a less mediated access to the source utterance, which allows them to better exploit speech information (e.g. prosody) without loss of information.

The paucity of parallel (audio, translation) data for direct ST has been previously addressed in different ways, ranging from 
\textit{model pre-training} to exploit knowledge transfer from ASR and/or MT \cite{berard2018end,bansal-etal-2019-pre,alinejad-sarkar-2020-effectively}\spapi{,}
\textit{knowledge distillation} \cite{Liu2019,gaido-2020-on-knowledge}, \textit{data augmentation}  \cite{jia2018leveraging,bahar-2019-specaugment,nguyen2019improving}, and  \textit{multi-task learning} \cite{weiss2017sequence,anastasopoulos-2018-multitask,bahar_2019,gaido-etal-2020-end}.
Thanks to these studies, the gap between the strong cascade models and the new end-to-end ones has gradually reduced during the last few years. 
As highlighted by the IWSLT 2020 Offline Speech Translation challenge results \cite{ansari-etal-2020-findings}, the rapid evolution of the direct approach has eventually led it to performance scores that are similar to those of cascade architectures. In light of this positive trend, we decided to adopt only the direct approach 
(described in Section \ref{sec:model}) 
for our participation in the 2021 round of the offline ST task.

Another interesting finding from last year's campaign concerns the sensitivity of ST models to different segmentations of the input audio. 
The 2020 winning system \cite{potapczyk-przybysz-2020-srpols} shows that, with a custom segmentation of the test data, the same model improved 
by 3.81 BLEU points the score achieved when using the basic segmentation  provided by the task organizers. 
This noticeable difference is due to a well-known problem in MT, ST and in machine learning at large: any mismatch between training and test data (in terms of domain, text style or a variety of other aspects) can cause unpredictable, often large, performance drops at test time. 
In ST, this is a critical issue, inherent to the nature of the available resources: while systems are usually trained on corpora that are manually segmented at sentence level, test data come in the form of unsegmented continuous speech.

A possible solution to this problem is to automatically segment the test data with a Voice Activity Detection (VAD) tool \cite{sohn_vad}. This strategy tries to mimic the sentence-based segmentation observed in the training data using pauses as an indirect (hence known to be sub-optimal) cue for sentence boundaries.
Custom segmentation strategies, which are allowed to IWSLT participants, typically go in this direction with the aim to reduce the data mismatch by working on evaluation data.
An opposite way to look at the problem is to work on the training data. In this case, the goal is to ``robustify'' the ST model to noisy inputs (i.e. sub-optimal segmentations) at training time, by exposing it to perturbed data where sentence-like boundaries are not guaranteed.
Our participation in the offline ST task exploits both solutions (see Section \ref{sec:mismatch}): at training time, by fine-tuning the model with a random segmentation of the available in-domain data; at test time, by feeding it with a custom hybrid segmentation of the evaluation data.

In a nutshell, our participation can be summarized as follows.
After a preliminary model selection phase that was carried out in order to select the best architecture, we adopted a pipeline consisting of: \textit{i)} ASR pre-training, \textit{ii)} ST training with knowledge distillation with an MT teacher, and \textit{iii)} two-step fine-tuning by varying the type and the amount of data between the two steps.
The second fine-tuning step, which was carried out on artificially perturbed data to increase model robustness, represents the main aspect characterizing our participation to this year's round of the offline ST task together with our custom automatic segmentation of the test set (see Section \ref{sec:mismatch}).
Our experimental results proved the effectiveness of our solutions: compared to a standard ST model and a baseline VAD-based method, on the MuST-C v2 English-German test set \cite{Cattoni2020mustc-v2}, the gap with optimal manual segmentation is reduced from 8.3 to 1.4 BLEU.

\section{Training data}
\label{sec:data}
To build our models, we used most of the training data allowed for participation.\footnote{\url{https://iwslt.org/2021/offline}}
They include: MT corpora (English-German text pairs), ASR corpora (English audios and their corresponding transcripts) and ST corpora (English audios with corresponding English transcripts and German translations).

\paragraph{MT.} Among all the available datasets, we selected those allowed for WMT 2019 \cite{barrault-etal-2019-findings} and OpenSubtitles2018 \cite{lison-tiedemann-2016-opensubtitles2016}. 
Some pre-processing was required to isolate and remove different types of potentially harmful noise present in the data. 
These include non-unicode characters, both on the source and target side of the parallel sentence pairs, which would have led to an increased dictionary size  hindering model training, and whole non-German target sentences (mostly in English). 
The cleaning of this two types of noise,
which was respectively performed using a custom script and Modern MT \cite{5d3fd94142224f40867fa83f7f447ee6}, resulted in the removal of roughly 25\% of the data, with a final dataset of $\sim$49 million sentence pairs.

\paragraph{ASR.} ASR corpora, together with the ST ones described below, were collected for the ASR training. In detail, the allowed native ASR datasets are: LibriSpeech \cite{7178964}, TEDLIUM v3 \cite{DBLP:conf/specom/HernandezNGTE18} and Mozilla Common Voice.\footnote{\url{https://commonvoice.mozilla.org/en/datasets}} In all of them, English texts were lowercased and punctuation was removed.

\paragraph{ST.} 
The ST benchmarks we used are essentially three:
\textit{i) }Europarl-ST (obtained from  European Parliament  debates -- \citealt{jairsan2020a}), \textit{ii) }MuST-C v2 (built from TED talks -- \citealt{Cattoni2020mustc-v2}), and \textit{iii)} CoVoST 2 (containing the translations of a portion of the Mozilla Common Voice dataset -- \citealt{wang-etal-2020-covost}).
To cope with the scarcity of ST data, we complemented these native ST corpora with synthetic
data.
To this aim, we used the MT system trained on the available MT data to translate into German the English transcripts of the aforementioned ASR datasets. The resulting texts were used as reference material during the ST model training.
The combination of native and generated data resulted in a total of about 1.26 million samples. The transcription-translation pairs were tokenized using, respectively, source/target-language SentencePiece \cite{sennrich-etal-2016-neural} unigram models trained on the MT corpora with a vocabulary size of 32k tokens. Similar to our last year's IWSLT submission \cite{gaido-etal-2020-end}, the entire dataset was used for training in a multi-domain fashion, where the two domains were \emph{native} (original ST data) and \emph{generated} (synthetic data).

Prior to the extraction of the speech features, the audio was pre-processed with the SpecAugment \cite{Park2019} data augmentation technique, which masks consecutive portions of the input both in frequency and in time dimensions.
From all the audio files, 80 log Mel-filter banks features were extracted using PyKaldi \cite{pykaldi}, filtering out those samples containing more than 3,000 frames. 
Finally, we applied utterance level Cepstral Mean and Variance Normalization both during ASR pre-training and ST training phases. The configuration parameters used are the default ones as set in \cite{wang2020fairseqs2t}.

\section{Model and training}
\label{sec:model}
In order to select the best performing architecture, we trained several Transformer-based models \cite{NIPS2017_3f5ee243}, 
which consist of 12 encoder layers, 6 decoder layers, 8 attention heads, 512 features for the attention layers and 2,048 hidden units in the feed-forward layers. 
The ASR and ST models are based on a custom version of the model by \cite{wang2020fairseqs2t}, which is a Transformer whose encoder has two initial 1D convolutional layers with \textit{gelu} activation functions \cite{hendrycks2020gaussian}.
Also, the encoder self-attentions were biased using a logarithmic distance penalty in favor of the local context as per \cite{Gangi2019}.
A Connectionist Temporal Classification (CTC) scoring function was applied as described in \cite{gaido-etal-2020-end}.
This was done by adding a linear layer to either the 6th, 8th or 10th encoder layer to map the encoder states to the vocabulary size and compute the CTC loss. 
The choice of the final architecture, depending on where the CTC loss is applied, was made based on sacreBLEU score \cite{post-2018-call} after training the models on MuST-C v1 En-De 
\cite{Cattoni2020mustc-v2}.
ST results computed on the test set are reported on Table \ref{tab:model-selection}. As it can be seen from the table, two models  obtained the highest, identical BLEU score (21.21): they both use logarithmic distance penalty but apply CTC loss to the 6th or the 8th encoder layer.

\begin{table*}
\centering
\begin{tabular}{lcc|c}
\hline
architecture & CTC encoder layer & distance penalty & \textbf{BLEU} \\
\hline
2d convolutional & 6 & no & 19.04 \\
\hline
1d convolutional & 6 & no & 21.16 \\
1d convolutional & 6 & log & 21.21 \\
\hline
1d convolutional & 8 & log & 21.21 \\
1d convolutional & 10 & log & 21.08 \\

\hline
\end{tabular}
\caption{\label{tab:model-selection}
Results of 1d convolutional architectures trained computing CTC loss at different layers and with/without distance penalty. Also the result of a 2d convolutional architecture is reported where the structure is exactly the same except for the use of a different type of convolution.
}
\end{table*}

\subsection{Training pipeline}
\label{subsec:pipe}
In the following, we describe the pipeline used to build our ST models, as anticipated in Section \ref{sec:intro}. In details, the ASR model is trained and its encoder used as starting point for the ST model, which is first trained via knowledge distillation and then fine-tuned on native and synthetic data. Then, a second fine-tuning step is performed on a perturbed version of a subset of the native data, focused on reducing the model performance drop over different segmentations.
For the initial ST training, we optimized KL divergence \cite{kullback1951} and CTC losses. For the first fine-tuning step, we optimized label smoothed cross entropy (LSCE) or CTC+LSCE while, for the second fine-tuning step, the models were refined using LSCE only, with a lower learning rate in order not to override
the knowledge acquired during the previous phases.

\paragraph{ASR pre-training.}
Due to the identical BLEU score obtained  by applying the CTC loss to the 6th and 8th layer during the ST model selection phase, we opted for training the ASR system using both these architectures, and selected the final model by looking at the Word Error Rate (WER) achieved by averaging 7 checkpoints around the best one. As shown in Table \ref{tab:ASR-results}, the best overall performing architecture is the one where the CTC is applied to the 8th encoder layer. Accordingly, we used this architecture to perform all the successive training phases.

\begin{table}
\centering
\begin{tabular}{l|r|r}
model & dev & test\\ 
\hline
CTC on 6th encoder layer & 8.67 & 12.19 \\
\textbf{CTC on 8th encoder layer} & \textbf{7.52} & \textbf{10.70} \\
\end{tabular}
\caption{\label{tab:ASR-results} 
Results of ASR pre-training in terms of WER. The dev and test sets used are, respectively, dev and tst-COMMON of MuST-C v1 En-De.}
\end{table}

\paragraph{Training with knowledge distillation.}
Two ST models, one with 12 and one with 15 encoder layers, were trained by loading the pre-trained ASR encoder weights and applying word-level Knowledge Distillation (KD) as in \cite{kim-rush-2016-sequence}.
In KD, a \emph{student} model is trained with the goal of learning how to produce the same output distribution as a \emph{teacher} model, and this is obtained by computing the KL divergence between the two output distributions. In our setting, the student and the teacher are respectively the ST system and an MT system that we trained on the MT data described in Section \ref{sec:data}.
It consists in a plain Transformer model with 6 layers for both the encoder and the decoder, 16 attention heads, 1,024 features for the attention layers and 4,096 hidden units in the feed-forward layers. 
Evaluated on the MuST-C v2 En-De test set, it achieved a BLEU score of 33.3.
For ST training with KD, we extracted only the top 8 tokens from the teacher distribution. According to \cite{tan2018multilingual}, this choice results in a significant reduction of the memory required, with no loss in final performance. At the end of this phase, we decided to keep the model with 15 encoder layers as it performs better than the one with 12 encoder layers by 1 BLEU point.

\paragraph{Fine-tuning step \#1: using  native and synthetic data.}
Once the KD training phase was concluded, we performed a multi-domain fine-tuning where the ST model was jointly trained on native and synthetic data optimizing LSCE or its combination with the CTC loss. 

\label{sec:results}
\begin{table*}[ht]
\centering
\begin{tabular}{l|ccccc|cccc}
\hline
model & MuST-C2 & \multicolumn{2}{c}{MuST-C2} & \multicolumn{2}{c|}{MuST-C2} & \multicolumn{2}{c}{IWSLT2015} & \multicolumn{2}{c}{IWSLT2015} \\
 & manual & \multicolumn{2}{c}{VAD (WebRTC)} & \multicolumn{2}{c|}{hybrid} & \multicolumn{2}{c}{VAD (LIUM)} & \multicolumn{2}{c}{hybrid} \\
\hline
1-FT LSCE & 27.6 & 20.8  & & 24.8 & & 16.1 & & 21.9 & \\
2-FT LSCE & - & 23.4 & (+2.6) & \textbf{26.4} & (+1.6) & 20.7 & (+4.6) & 22.7 & (+0.8) \\
\hline
1-FT LSCE+CTC & 27.7 & 19.9 & & 25.3 & & 14.0 & & 21.7 & \\
2-FT LSCE+CTC & - & \textbf{23.7} & (+3.8) & 26.3 & (+1.0) & \textbf{20.9} & (+6.9) & \textbf{23.1} & (+1.4) \\
\hline
\end{tabular}
\caption{\label{tab:final_results}
Results of the best architectures deriving from KD training after one or two fine-tuning steps. 1-FT stands for one-step fine-tuning and 2-FT stands for two-step fine-tuning (see Section \ref{sec:model}). MuST-C v2 results on manual segmentation have been not computed for the 2-step fine-tuned models as we were interested in the evaluation of the improvement on automatically segmented data.
}
\end{table*}

\section{Coping with training/test data mismatch}
\label{sec:mismatch}
As mentioned in Section \ref{sec:intro}, the segmentation of audio files is a crucial aspect in ST. In fact, mismatches between the manual segmentation of the training data and the automatic one required when processing the unsegmented test set can produce significant performance drops.
To mitigate this risk, we worked on two complementary fronts: at training and inference time.
At training time, we tried to robustify our model by fine-tuning it on a randomly segmented subset of the training data.  
At inference time, we applied an automatic segmentation procedure to the test set in order to feed the model with input resembling, as much as possible, the gold manual segmentation.
These two solutions, which characterize our final submission, are explained in the following.

\paragraph{Fine-tuning step \#2: using randomly segmented data.} 
For the second fine-tuning step, we re-segmented the MuST-C v2 En-De training set following the procedure described in \cite{gaido2020contextualized}. The method consists in choosing a random word in the transcript of each sample, and using it as sentence boundary instead of the linguistically-motivated (sentence-level) splits provided in the original data. The corresponding audio segments are then obtained by means of  audio-text alignments performed with Gentle.\footnote{\url{https://github.com/lowerquality/gentle/}} Similarly, the German translation of each re-segmented transcript is extracted with cross-lingual alignments generated by a fast\_align  \cite{dyer-etal-2013-simple} model trained on all the MT data available for the task and on MuST-C v2. In case either of the alignments is not possible (because fast\_align is not able to align enough words or Gentle does not recognize the position of the word in the audio), the sentence is discarded. 
The resulting material, which  contains $\sim5\%$ less segments than the original MuST-C release, was then used for our second (and final) fine-tuning step.   
As already stated, we used only the LSCE loss for this stage.

\paragraph{Automatic segmentation of the test data.}
At inference time, the test set was segmented with an hybrid approach that considers both the audio content and the length of the resulting segment \cite{gaido2021voice}.
Specifically, every segment is ensured to be at least 17s and at most 20s long, but the exact splitting position is determined by the longest pause detected within this interval. Pauses are identified with the WebRTC VAD tool \cite{10.5555/2432294}, using 20ms as \textit{frame duration} and 2 as \textit{aggressivity} level.

\section{Experimental settings}
Our implementation is built on top of fairseq Pytorch library  \cite{ott2019fairseq}. All our models were trained using the Adam optimizer \cite{DBLP:journals/corr/KingmaB14} with
$\beta_1=0.9$, $\beta_2=0.98$. During 
training, the learning rate was set to increase linearly from 0 to 2e-3 for the first 10,000 warm-up steps and then to decay with an inverse square root policy.
Differently, the learning rate was kept constant for model
fine-tuning, with a value of 1e-3 for the first fine-tuning step and 1e-4 for the second one.

All the trainings were performed on 2 Tesla V100 GPUs with 32GB RAM. We set the maximum number of tokens to 10k per batch and 8 as update frequency. For generation, the maximum number of tokens was increased to 50k, using a single Tesla V100 GPU and by applying a standard 5-beam search strategy.

\section{Results}
For the evaluation of the fine-tuned models we considered three different test sets: MuST-C v2 En-De tst-COMMON, IWSLT 2015 and 2019 test sets (available on the Offline ST task Evaluation Campaign web page\footnote{\url{https://iwslt.org/2021/offline}}). While for MuST-C v2 we originally had a manual segmentation of the audio files, for the IWSLT 2015 and 2019 test sets the organizers provided only automatic segmentations obtained by the LIUM VAD tool 
\cite{meignier:hal-01433518}.
Furthermore, we segmented MuST-C v2 tst-COMMON using the WebRTC VAD tool to have a comparable framework. 
Table \ref{tab:final_results} reports the results before and after the second fine-tuning step, which clearly show that performing the additional training on randomly segmented data  highly improves the performance in the non-manual segmentation case, by up to 6 BLEU points.
We also created an ensemble with the best two models reported in Table \ref{tab:final_results}, whose KD training also used CTC loss. Results are not reported here since ensembling did not bring any improvement in terms of BLEU score compared to the two separate models. 
A possible motivation is
that our two-step fine-tuning process is already sufficient to build a robust model, which is capable of generalizing without the need of combining two or more model outputs. 

For our \emph{primary} submission, we chose the two-step fine-tuned model that uses the LSCE+CTC losses for the first fine-tuning step (2-FT LSCE+CTC) since it achieved the highest BLEU on automatically segmented data.
In order to measure the contribution of 
fine-tuning on randomly segmented data also on the official evaluation set, we selected the same model before the second fine-tuning step (1-FT LSCE+CTC) as our \emph{contrastive} submission.

Our primary submission scored 30.6 BLEU on the tst2021 test set considering both references while our contrastive scored 29.3 BLEU, showing the effectiveness of our fine-tuning step. 
In addition, our primary submission scored 24.7 BLEU on the tst2020 test set.

\section{Conclusions}
We described FBK’s participation in the IWSLT2021 Offline Speech Translation task \cite{iwslt:2021}. Our work focused on a multi-step training pipeline involving data augmentation (SpecAugment and MT-based synthetic data), multi-domain transfer learning (KD training first and then fine-tuning on synthetic and native data) and ad-hoc fine-tuning on randomly segmented data.
Based on the experimental results, our submission was characterized by the use of the CTC loss on transcripts during word-level knowledge distillation training, followed by a two-stage fine-tuning aimed to fill the gap between the performance of models when tested on manual and automatically segmented data. This huge gap was pointed out in our last year submission \cite{gaido-etal-2020-end}, where we highlighted that some strategies should have been adopted in order to mitigate the problem.
This paper demonstrates that, following the above-mentioned pipeline, together with some data-driven techniques, we can obtain significant improvements in the performance of end-to-end ST systems. Research in this direction will help us to build models that are not only competitive with cascaded solutions, but also able to handle different segmentation strategies which are going to be more frequently used in the future.

\bibliographystyle{acl_natbib}
\bibliography{acl2021}


\end{document}